\documentclass{article}
\pdfoutput=1
\usepackage{ijcai22}

\usepackage{times}

\usepackage{soul}
\usepackage{url}
\usepackage[hidelinks]{hyperref}
\usepackage[utf8]{inputenc}
\usepackage[small]{caption}
\usepackage{graphicx}
\usepackage{amsmath}
\usepackage{booktabs}
\usepackage{color}
\title{An Efficient Industrial Federated Learning Framework for AIoT: \\ A Face Recognition Application}

\author{
Youlong Ding$^1$\footnote{Contact Author}\and
Xueyang Wu$^2$\and
Zhitao Li $^1$\and
Zeheng Wu$^{3}$\and
Shengqi Tan$^3$\and\\
Qian Xu$^{2,3}$\and
Weike Pan$^{1}$\And
Qiang Yang$^{2,3}$\\
\affiliations
$^1$Shenzhen University, Shenzhen, China\\
$^2$The Hong Kong University of Science and Technology, Hong Kong SAR, China\\
$^3$WeBank Inc., Shenzhen, China\\
\emails
dingyoulon@gmail.com,
xwuba@connect.ust.hk,
lizhitao2018@email.szu.edu.cn,\\
\{frankwu, elontan\}@webank.com,
qianxu@ust.hk, panweike@szu.edu.cn, qyang@cse.ust.hk
}

\begin{document}

\maketitle

\begin{abstract}
Recently, the artificial intelligence of things (AIoT) has been gaining increasing attention, with an intriguing vision of providing highly intelligent services through the network connection of things, leading to an advanced AI-driven ecology. However, recent regulatory restrictions on data privacy preclude uploading sensitive local data to data centers and utilizing them in a centralized approach. Directly applying federated learning algorithms in this scenario could hardly meet the industrial requirements of both efficiency and accuracy. Therefore, we propose an efficient industrial federated learning framework for AIoT in terms of a face recognition application. Specifically, we propose to utilize the concept of transfer learning to speed up federated training on devices and further present a novel design of a private projector that helps protect shared gradients without incurring additional memory consumption or computational cost. Empirical studies on a private Asian face dataset show that our approach can achieve high recognition accuracy in only 20 communication rounds, demonstrating its effectiveness in prediction and its efficiency in training.


\end{abstract}

\section{Introduction}
With the rapid growth of smart devices and high-speed networks, the internet of things (IoT) has been gaining more and more attention. The IoT is a network that connects a large number of end devices equipped with sensors~\cite{atzori2010internet,khan2018iot},
which collects multiview and multimodal data from the network and allows further comprehensive analysis. 


Recently, many researchers and engineers have been making efforts to empower end devices in the IoT with artificial intelligence (AI) techniques to form a network of AI agents, dubbed artificial intelligence of things (AIoT)~\cite{luo2018aiot}. The vision of AIoT is to not only connect the data around the network but also provide highly intelligent services, which leads to an advanced  AI-driven ecology. 

Thanks to the recent advances in deep learning and hardware, AIoT has been rapidly developing. On the one hand,  deep learning models are becoming more and more lightweight due to the advances in model compression~\cite{han2015deep}. On the other hand, the devices provide additional hardware acceleration using GPUs or NPUs to support the AI applications. These two trends accelerate the industrial deployment of AIoT.  

Deep learning algorithms, however, are data hungry. Model training benefits a lot from a large and representative dataset, such as ImageNet~\cite{5206848} and COCO~\cite{lin2014microsoft}. It is well known that the more training data, the better performance a model can achieve. However, data are usually highly scattered among different AIoT end devices (i.e., devices with sensors that directly collect data) in industrial scenarios. Meanwhile, since industrial data are typically proprietary and sensitive, regulations, such as the newly enforced European Union General Data Protection Regulation (GDPR)~\cite{voss2016european}, preclude uploading them to data centers and being utilized in a centralized manner. The risk of privacy leakage poses new challenges to the practical deployment of AIoT.

A straightforward solution to comply with the need for collaborative learning and data protection regulations is to utilize the federated learning (FL) technique, a distributed architecture in which a central server coordinates a fleet of clients to collaboratively train a machine learning model without sharing their local data~\cite{yang2019federated}.
One of the most widely used federated learning algorithms is FedAvg~\cite{mcmahan2017communication}, in which the updated local models of the clients are transferred to the server, which further aggregates the local models by averaging the parameters to update the global model, and the process iterates until convergence. 

However, due to the small memory size and limited computing power of AIoT devices, it is challenging to apply federated learning algorithms to this scenario directly. This difficulty comes from the following two aspects. Firstly, the non-convexity of deep learning objectives increases the required convergence time. For example, when applying a commonly used optimizing strategy, stochastic gradient descent (SGD) ~\cite{bottou2010large}, to the objective, it typically needs hundreds or thousands of epochs until convergence, which is unaffordable due to the limited network bandwidth and transmitting speed. Secondly, to avoid privacy leakage risk from the shared gradients of each client, differentially private algorithms are incorporated into the training process by adding calibrated noise into model gradients~\cite{DBLP:conf/ccs/AbadiCGMMT016}, which protects the shared gradients with rigorous guarantees at the cost of efficiency degradation.  

To address the challenges mentioned above, we propose an efficient industrial federated learning framework for AIoT. We demonstrate our framework in terms of a face recognition application while it paves the way for a broader scope of applications in this scenario. 

Inspired by the concept of transfer learning~\cite{pan2009survey}, we propose to first pre-train a face representation model on a publicly available dataset, where the trained parameters will serve as the initialization of the global model for FedAvg. This approach can significantly reduce the required workload of end devices since they only need to slightly adjust model parameters (or fine-tune them) to fit local data collaboratively. The central server has sufficient computing resources to undertake the heavy computational task to provide a good initialization, lifting the burden on AIoT devices. For the second challenge, we further present a novel design of a local projector, which prevents attackers from performing full-forward propagation and hopefully serves as an alternative to differentially private algorithms in this scenario. In addition, we present several empirical strategies for effective federated fine-tuning, which allows us to achieve nearly lossless prediction performance compared with the one in centralized settings. We validate our proposed framework on a private Asian face dataset. The empirical results show that our approach can achieve high prediction accuracy in only 20 communication rounds, which demonstrates its effectiveness and efficiency.

\begin{figure*}[!htb]
    \centering
    \includegraphics[width=0.9\textwidth]{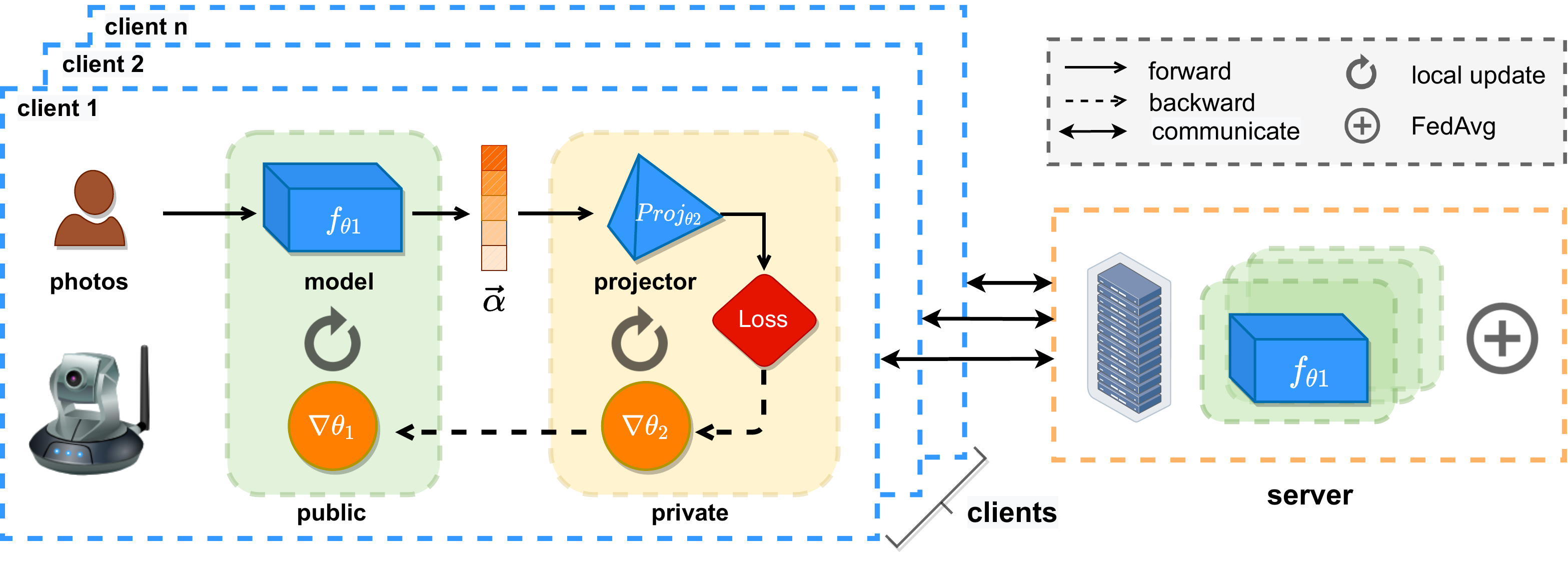}\\
    \caption{Illustration of our efficient federated training framework for AIoT devices. The server will first send a pre-trained model to each client, which further appends an additional private projector to the representation model. Only the public module will be shared and uploaded to the central server in each communication round for aggregation. Privacy protection is achieved by blocking the full-forward propagation for the attacker due to the limited access to the full model (lack of local projector).}
    \label{fig:model}
\end{figure*}

\section{Related Work}
\subsection{Artificial Intelligence of Things}
With the aid of the fifth-generation (5G) wireless communication and AI technology, numerous sensors can form an AIoT system with a cloud computing server to collect, store, process, and analyze data, as well as control the system~\cite{dong2021technology}. \cite{zhang2021aiot} proposed an AIoT-based system for real-time monitoring of tunnel construction. \cite{lai2020cognitive} proposed an expected advantage learning method for applying reinforcement learning algorithms to mechanical control of AIoT. \cite{sun2021artificial} presented a digital-twin-based virtual shop to provide the users with real-time feedback about the details of the product by leveraging AIoT analytics. However, current existing works all assume that the collected data by AIoT devices can be freely transmitted to the server for processing and analysis, which will cause privacy issues.

\subsection{Federated Learning}
A popular federated learning algorithm is \texttt{FedAvg}~\cite{mcmahan2017communication}. There are four steps in one round of \texttt{FedAvg}. First, the server sends an initial model to clients. Second, the clients update the parameters of model based on their local data. Third, the local models (or equivalently, the local gradients) are sent to the central server. Finally, the server averages the model weights to produce a global model for the training in the next round. 
A key challenge in federated learning is the heterogeneity of data distribution on different parties~\cite{kairouz2021advances}. The main idea to tackle this problem is to regularize local models during the local training step. \texttt{FedProx}~\cite{li2020federated-fedprox} directly limits the local updates by the $\ell_2$-norm distance. \texttt{SCAFFOLD}~\cite{karimireddy2020scaffold} corrects the local updates via variance reduction. \texttt{MOON}~\cite{li2021model} utilizes the similarity between model representations to correct the local training by conducting contrastive learning in the model level. In this paper, we mainly use \texttt{FedAvg} and other alternatives can serve as a drop-in replacement of \texttt{FedAvg} in our framework as required.

Federated transfer learning (FTL)~\cite{yang2019federated,ftlsurvey} is a noteworthy variant of the federated learning, which allows knowledge to be transferred across domains that do not have many overlapping features and users, and training with heterogeneous data across participants may present additional challenges compared to FL. To overcome these challenges,~\cite{dimitriadis2020federated} presented a dynamic gradient aggregation method that weights the local gradients during the aggregation step. Note that even though our federated learning framework takes inspiration from the concept of transfer learning, our method does not belong to the class of federated transfer learning, since the feature space of the data is identical across all participants.

\subsection{Face Recognition}
Face recognition systems\cite{turk1991eigenfaces} can be applied to a wide variety of situations, including criminal identification, user verification, and human-computer interaction. The common setting in face recognition is open-set protocol, where the test identities are often disjoint from the training set. Under this circumstance, more discriminative features are needed to be extracted from the neural networks to perform accurate prediction. In other words, we should encourage the intra-class compactness and inter-class separability, which means the maximal intra-class's distance is expected to be smaller than minimal inter-class distance under a certain metric space. Center loss~\cite{wen2016discriminative} simultaneously learns a center for deep features of each class and penalizes the distances between the deep features and their corresponding class centers. SphereFace\cite{Liu_2017_CVPR} employs an angular softmax loss that enables CNNs to learn angularly discriminative features. ArcFace~\cite{deng2019arcface} adds an additive angular margin loss to obtain highly discriminative features for face recognition, achieving the state-of-the-art performance on a wide variety of datasets. 





\section{Methodology}
In this section, we are going to discuss our main methodology for conducting federated learning on AIoT devices, which allows us to efficiently train a face recognition system without sharing face images collected on end devices.

In the beginning, the server will first train a model on a public face dataset and use the trained parameters as the initialization for \texttt{FedAvg}, which aims at speeding up the convergence for training on AIoT devices, inspired by the concept of transfer learning~\cite{pan2009survey}. This process is accomplished on computer clusters within the server.
Then the pre-trained face model, which consists of several convolution layers, collectively denoted by $f_{\theta_1}$, is sent to each end device. Upon receiving it, each device will append to the pre-trained model an additional private projector, denoted by $Proj_{\theta_2}$, which is a revised classification layer based on ArcFace. Thus, the full parameter is denoted by $\theta=[\theta_1, \theta_2]$. Given an input image $X$, the forward propagation is defined as:
\begin{align}
    &\vec{\mathbf{\alpha}} = f_{\theta_1}(X), ~~~\hat{o} = Proj_{\theta_2}(\vec{\alpha})
\end{align}
where $\vec{\alpha}$ is the latent representation of the input face, and $\hat{o}$ is the prediction logits used for evaluating cross-entropy loss. The appended layer $Proj_{\theta_2}$ is randomly initialized by each device. When the training process starts, each client adjusts the parameters of the whole model based on the local training data to minimize the local objective, which involves jointly fine-tuning the pre-trained parameters and training the newly added parameters from scratch. The gradients conditioned on the input image $X$ and its label $y$ are calculated via backpropagation procedure as follows,
\begin{align}
    &\nabla_{\theta_2}(X, y) = \frac{\partial l(\hat{o}, y)}{\partial \theta_2} = \frac{\partial l(\hat{o}, y)}{\partial \hat{o}}\cdot \frac{\partial \hat{o}}{\partial \theta_2}\\
    &\nabla_{\theta_1}(X, y) = \frac{\partial l(\hat{o}, y)}{\partial \theta_1} = \frac{\partial l(\hat{o}, y)}{\partial \hat{o}}\cdot \frac{\partial \hat{o}}{\partial \vec{\alpha}} \cdot \frac{\partial \vec{\alpha}}{\partial \theta_1}
\end{align}
The pre-trained module, $f_{\theta_1}$, which we have updated during the local training, will be sent back to the server for averaging, whereas the new module, $Proj_{\theta_2}$, never leaves the device. 

This treatment is the key design of our approach to protecting privacy. Specifically, the whole model is now composed of a public module (i.e., $f_{\theta_1}$) and a private module (i.e., $Proj_{\theta_2}$), and only the gradients (or model parameters) of the public module will be uploaded to the server for averaging. The standard process of model inversion~\cite{zhu2019deep,zhao2020idlg,geiping2020inverting} attempts to recover the input data from the shared gradients. Given full model parameters $F_{\theta}$ along with shared gradients $\nabla_{\theta}$, this attack can be generally formulated as optimizing the following objective,
\begin{equation}
    J(\tilde{X}, \tilde{y}) = \left\|\nabla_{\theta}(X, y) - \frac{\partial l(F_{\theta}(\tilde{X}), \tilde{y})}{\partial \theta}\right\|^2
\end{equation}
where $\nabla_{\theta}(X, y)$ is the model gradients from one client, and $\tilde{X}, \tilde{y}$ is the recovered data from the shared gradients by the attacker.
Our protection comes from blocking the full-forward propagation. Since only part of the full model $F_{\theta}$ is public (i.e., $f_{\theta_1}$), the attacker could not evaluate the value of $F_{\theta}(\tilde{X})$ without the private part $Proj_{\theta_2}$ for forward propagation, and therefore the corresponding gradients for $\tilde{X}, \tilde{y}$, i.e., $\frac{\partial l(F_{\theta}(\tilde{X}), \tilde{y})}{\partial \theta}$, becomes intractable.

This novel design reflects a meaningful insight into the deep face model, where the stacked layers before the final output layer can be viewed as extracting a compact representation of the input face image. And feeding that representation into the top layer is equivalent to jointly training a linear classification model which takes that representation as input. Therefore, our approach can be understood from another perspective. Note that the goal of a face model is to represent human faces in a low-dimensional manifold embedded in the original high-dimensional data space, which means that $f_{\theta_1}$ is what we are really concerned with. The incorporation of the classification layer $Proj_{\theta_2}$ serves as the projector that helps us evaluate the quality of extracted features and therefore enables the supervised training of $f_{\theta_1}$. Since our goal through federation is to collaboratively train a powerful face feature extraction model, the direct application of \texttt{FedAvg} on the parameters of that model, i.e., $f_{\theta_1}$, yields the same treatment as our method. 

When the training procedure is finished, each local device will simply discard $Proj_{\theta_2}$ and use $f_{\theta_1}$ for local inference. The prediction is given by comparing the taken face image to the images of all people in oracle, stored in the form of latent representations $\{\vec{\alpha}_i, i\in \mathcal{P}\}$, where $\mathcal{P}$ denotes the set of people of interest. The comparison is based on the Euclidean distance in the latent representation space,
\begin{equation}
    \hat{y} = argmin_{i\in \mathcal{P}}\left\|f_{\theta_1}(X) - \vec{\alpha}_i\right\|_2
\end{equation}
This prediction rule gives us a nonlinear decision boundary within the latent representation space, in contrast to the linear boundary used for training. Note that $f_{\theta_1}$ is stored on devices, which enables low-latency inference, and contributes to privacy since inference does not require communicating local images to the central server.


\begin{figure*}[!htb]
    \centering
    \includegraphics[width=0.9\textwidth]{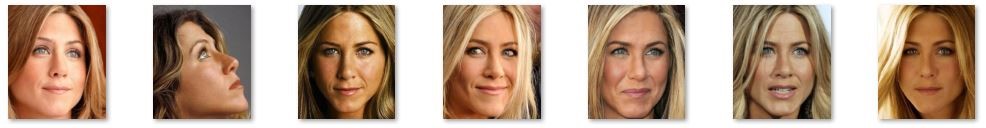}\\
    \caption{Face samples taken from the public MS-Celeb-1M dataset, which uses all the possibly collected face images of one million celebrities on the web. This dataset mainly consists of European and American face photos.}
    \label{fig:sample}
\end{figure*}

\section{Effective Federated Fine-tuning}
In this section, we present our strategies that help boost the performance of federated fine-tuning without incurring additional computational or communication costs. In federated learning under non-i.i.d. data distribution across participants, the local training phase may lead the model to learn a poor face representation due to the skewed local data distribution, known as a drift in local update. Our central goal through the proposed strategies is thus to migrate the effect of client bias during federated learning.

\subsection{Freeze BatchNorm Statistics}
To the best of our knowledge, we are the first to propose to freeze BatchNorm statistics during training. Batch normalization is widely used to normalize the inputs across mini-batches (i.e., zero-mean and unit-variance), which is beneficial for stabilizing and accelerating neural network training~\cite{ioffe2015batch}. The statistics used in batch normalization depend on other samples in the same mini-batch. Specifically, assuming the input is a vector $X$ which contains all features of a sample, the operation is defined as,
\begin{equation}
    \operatorname{BatchNorm}(X)=\boldsymbol{\gamma} \odot \frac{X-\mu}{\sqrt{\sigma^2 + \epsilon}} + \boldsymbol{\beta}
\end{equation}
where $\odot$ is the element-wise product (i.e., the Hadamard product), $\mu$ and $\sigma$ are the mean and variance of $X$, $\boldsymbol{\gamma}$ and $\boldsymbol{\beta}$ are learned scaling factors and bias terms.
Typically, $\mu$ and $\sigma$ are updated in every iteration as a running estimation of the statistics of the whole dataset. We, however, freeze the statistics (i.e., accumulated mean $\mu$ and variance $\sigma$) in the batch normalization layer of the model during fine-tuning. The intuition is that the pre-trained model has already seen a huge number of training samples, the statistics therefore should be more comprehensive than that obtained from the local data on an end device. In other words, the original statistics could properly reflect global data distribution and avoid performance deterioration due to local training, which motivates us to freeze them rather than updating them during local training.

\subsection{Warm-up Fine-tuning} 
Warm-up training~\cite{goyal2017accurate} is a useful strategy for training deep learning models, which reduces update variance and stabilizes the model at the beginning of training. In the warm-up heuristic, we use a small learning rate at the beginning and then switch back to the initial learning rate when the training process is stable. Here we employ a similar idea in federated fine-tuning, where the instability results from the attempt to adapt the pre-trained model to local data distribution. Specifically, we linearly increase our learning rate from zero to the initial value in the first several iterations, following the most common choice in warm-up training. Suppose the initial learning rate is $\eta$, we will set the learning rate as follows,
\begin{equation}
    \operatorname{learning\;rate} = \frac{i\eta}{m}, 1\leq i \leq m
\end{equation}
where $i$ denotes the iteration number and $m$ denotes the number of mini-batches we use for warm-up.



\section{Experiments}
\subsection{Datasets}
We use the public MS-Celeb-1M dataset~\cite{guo2016ms} for pretraining the model (Figure~\ref{fig:sample}). We locate and align each face, resizing each image to the shape of 112 $\times$ 112. We then perform data cleaning which finally results in a dataset consisting of 5,822,653 images for 85,742 distinct people.

\begin{figure}[!htb]
	\begin{center}
		\begin{tabular}{c}
		\includegraphics[width=200px]{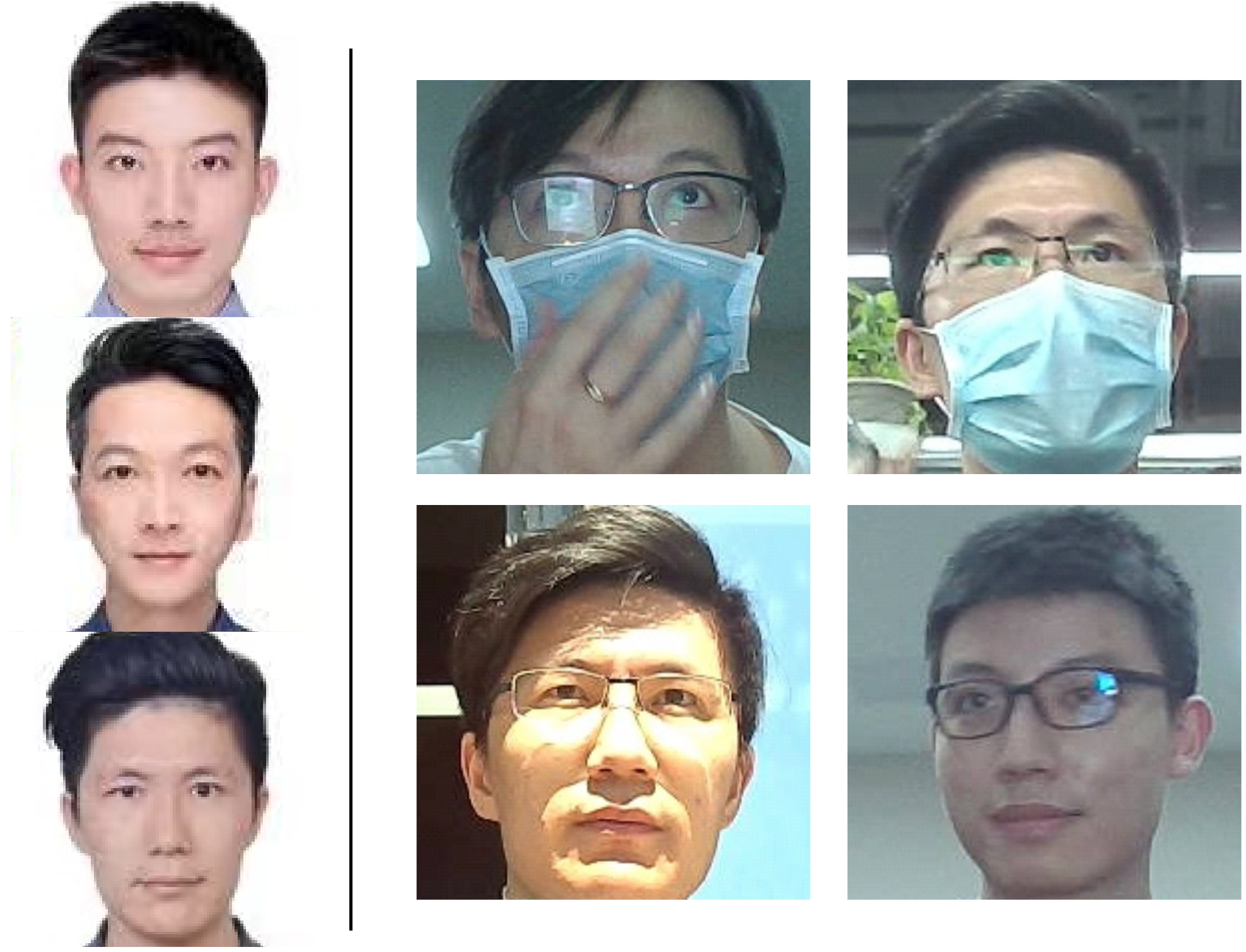}
		\end{tabular}
	\end{center}
	\caption{Face samples taken from the private dataset. Photos in the left side are taken under standard conditions, corresponding to the reference images in oracle, whereas photos in the right side are captured by some AIoT devices. Displayed photos have been authorized.}
	\label{fig:privatedata}
\end{figure}

For the local images on each device, we use a private dataset (Figure~\ref{fig:privatedata}) which consists of 20,551 face images of 4,998 East Asians. Each person has an image taken under standard conditions along with multiple labeled photos captured in the wild by cameras. The standard one is actually the reference photo stored in oracle, while others may have different illumination conditions or occlusions (i.e., mask). The images of the first 3000 people are used for training, and the images of the left 1998 people are used for test. The test set is in form of pairs of images. If the two images of a pair belong to the same person, the label is 1, otherwise 0.

Note that there exists a domain shift between the dataset for pre-training and the dataset for federated training. The public dataset, MS-Celeb-1M, mainly consists of European and American photos. The facial features of Asian people are quite different from that of European and American people, which motivates us to fine-tune the model on an in-domain dataset rather than directly deploy the model trained on the public dataset.

\begin{figure*}[!htb]
    \centering
    \includegraphics[width=0.9\textwidth]{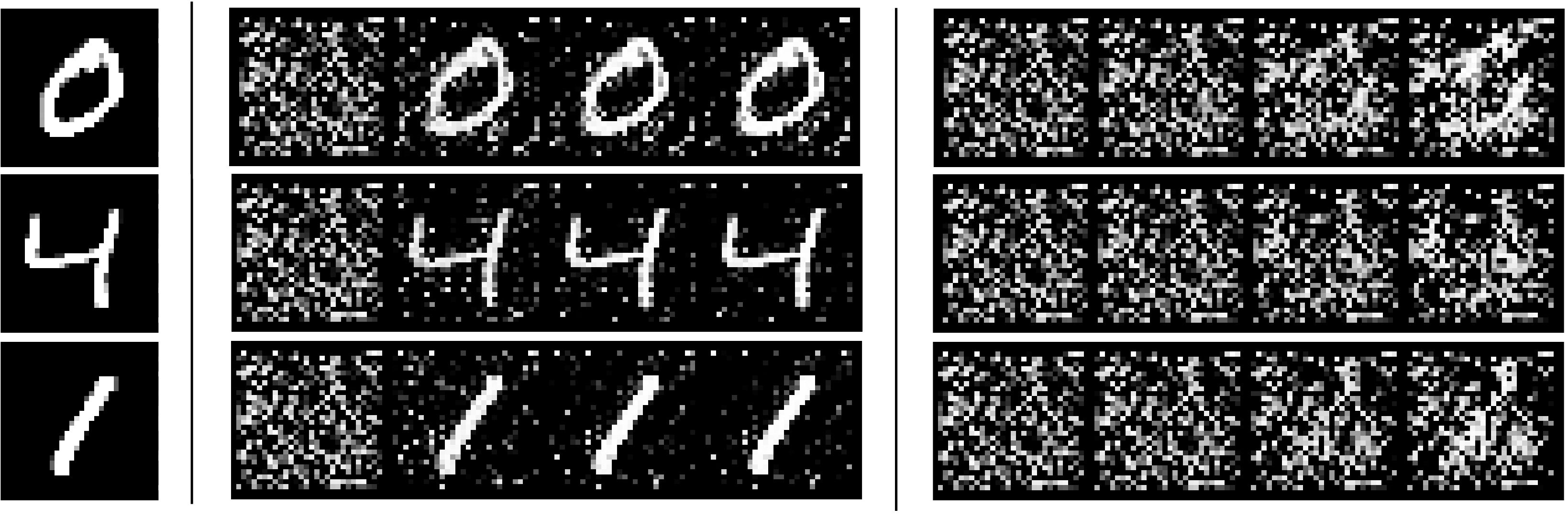}\\
    \caption{Qualitative evaluation of attack experiments. The left side images show the original input images for training. The processes of DLG attacks on the baseline and our approach are displayed in the middle and right, respectively.}
    \label{fig:attack}
\end{figure*}
\subsection{Experimental Settings}
For the face representation model $f_{\theta_1}$, we use the same network architecture as MobileFaceNet~\cite{chen2018mobilefacenets}, a lightweight and efficient backbone model suitable for AIoT,  which outputs a 512-dimensional latent representation vector $\vec{\alpha}$. For model pre-training on the server, we use the ArcFace layer as our classifier, with the scale hyper-parameter value of 64 and the margin hyper-parameter value of 0.5, following the standard settings for ArcFace. For training in the FL setup at each individual client, we implement the private projector with a new ArcFace layer.
We pre-train the model for 150 epochs in the server and conduct federated fine-tuning on local devices for only 20 communication rounds. The local epochs in one communication round are set to 1. We choose a learning rate of 0.001 to update the model locally on each of the clients. 

\subsection{Prediction Accuracy}
\textbf{Baselines.} We compare our approach with the following baselines.
\begin{itemize}
    \item \textbf{Centralized.} All local data on devices are collected to central server for fine-tuning.
    \item \textbf{Pre-trained.} We directly use the pre-train model for inference, without fine-tuning.
    \item \textbf{SOLO.} We fine-tune the pre-trained only with local data, do not conduct federated learning.
\end{itemize}

\noindent \textbf{Results.} We report our empirical results of accuracy in Table~\ref{tb:acc}. We evaluate the accuracy when the false acceptance rate (FAR) is set to 1e-3 and 1e-4, which measures the likelihood that the biometric security system will incorrectly accept an access attempt by an unauthorized user, defined by the ratio of the number of false acceptances divided by the number of identification attempts. Due to the domain shift mentioned before, the pre-trained model has poor performance on the test dataset. Our federated learning approach leads to an accuracy gain of over 8\% for FAR@1e-3 and around 4\% for FAR@1e-4. Note that these results are quite close to that obtained under the centralized learning setting, where all the training data on the local devices have been collected on the central server. The performance gap between federated learning (FL) and centralized learning (CL) is less than 1\%, which further demonstrates that our approach can achieve privacy protection only at a modest cost of model quality.

\begin{table}[!ht]
\begin{center}
\caption{Prediction performance of the baselines and our approach. We run five trials and report the mean and standard deviation.}
\label{tb:acc}
\begin{tabular}{cccc}
\toprule
Method & FAR@1e-3 $\uparrow$ &  FAR@1e-4 $\uparrow$\\ \midrule
Centralized & $73.58\% \pm 0.17\%$ & $64.26\% \pm 0.16\%$ \\\midrule
Pre-trained & $64.32\%$ & 59.68\%\\
SOLO & $67.15\% \pm 0.19\%$ & $61.34\% \pm 0.11\%$ \\
\textbf{Ours} & $\textbf{72.83\%} \pm 0.22\%$ & $\textbf{63.51\%} \pm 0.25\%$ \\\bottomrule
\end{tabular}
\end{center}
\end{table}

\subsection{Ablation Study}
To study the effect of each proposed strategy, we compare the performance of our approach with 1) training from scratch without starting at the pre-train model, 2) updating batch norm statistics as usual, and 3) finetuning without warm-up. The results are shown in Table~\ref{tb:ablation}. Note that the accuracy of the approach without pre-training is awkwardly zero, which means that if we train from scratch, the model is far from convergence under the same communication cost (20 communication rounds). The usual update of batch norm statistics leads to huge performance deterioration, only achieving about 50\% accuracy on both metrics, which indicates that the model's prediction is nearly random given any input image pair. Note that this accuracy is even worse than the performance of the pre-trained model, emphasizing the necessity of freezing batch norm statistics. Moreover, warm-up fine-tuning brings us around 1\% performance gain.
\begin{table}[!ht]
\centering
\caption{Prediction performance of the constructed ablation baselines and our approach. We run five trials and report the mean and standard deviation.}
\label{tb:ablation}

\begin{tabular}{ c c c c} 
\toprule
\multicolumn{1}{c}{Method} & FAR@1e-3 &  FAR@1e-4\\ \midrule
w/o pre-train & 0.00\% $\pm 0.00\%$ & 0.00\% $\pm 0.00\%$\\
w/o frozen BN & $51.68\% \pm 1.34\%$ & $50.29\% \pm 1.21\%$ \\
w/o warmup & $71.95\% \pm 0.32\%$ & $61.87\% \pm 0.49\%$ \\
\textbf{Ours} & $\textbf{72.83\%} \pm 0.22\%$ & $\textbf{63.51\%} \pm 0.25\%$ \\\bottomrule
\end{tabular}
\end{table}
\subsection{Attack Experiments}
We conduct attack experiments to validate the effectiveness of our approach in terms of privacy protection. It is worth noting that, for images of relatively high resolutions, conducting a successful attack turns out very challenging and may take plenty of time and computing resources even when we do not utilize any technique for protecting the gradients. In other words, it is inappropriate to attribute a failed attack to the superiority of our approach. Therefore, we choose MNIST~\cite{lecun1998gradient} in the experiments, which is a famous toy dataset of hand-written digits, making the results more convincing to demonstrate the effectiveness of our method.

\noindent\textbf{Threat Model.} In this work, we assume that all participants, including the clients and the server, are honest-but-curious adversaries. They possibly do some calculations to obtain the private data of clients by observing the model weights and updates, but they do not maliciously modify their inputs or parameters for the attack purpose. We will use a representative privacy attack~\cite{zhu2019deep} on our method. Note that in order to perform the full forward propagation for obtaining gradients with respect to the dummy input, the attacker must randomly initialize a projector and jointly optimize the input and the projector.

\noindent\textbf{Results.} Figure~\ref{fig:attack} shows the qualitative evaluation of the attack experiments, in which we select the first three data points of MNIST for display. The total iterations are set to 100 and further increasing that number will not bring us more benefits. We compare our method with the baseline, which does not utilize any technique for privacy protection. For the baseline, the original input images have been well recovered from the gradients after some iterations, indicating a successful attack. In contrast, for our approach, the attacker is hard to recover the original input. The recovered images have much noise and the strokes are not clear enough to be distinguished from the background. Note that in order to make the results more convincing, for our approach, we use the ground truth for the initialization of the projector, which should have been randomly initialized by the attacker. In other words, the displayed results of our approach are the attacker's upper bound in terms of the recovery quality, which further demonstrates the effectiveness of our method.

\subsection{Efficiency Analysis}
To demonstrate the efficiency of our method, we then study its memory/disk consumption and computational cost. The statistics are reported in Table~\ref{tb:efficiency}. Note that the model has only 1.20M parameters and about 35 MB total memory usage in training, which allows fitting the model into the on-chip SRAM cache rather than the off-chip DRAM memory of typical AIoT devices. For the computational cost, we evaluate the theoretical amount of floating-point arithmetics (FLOPs) and theoretical amount of multiply-adds (MAdd), both having a value below 0.5G. For inference, we need to store latent representations for all people, i.e., $\{\vec{\alpha}_i, i\in \mathcal{P}\}$, which takes up about 35MB on disk. We can see that the efficiency of our approach makes it suitable for deploying on real-world industrial AIoT devices.

\begin{table}[!ht]
\centering
\caption{Memory consumption and computational cost of our approach. The computational statistics are calculated for one iteration given a data point.}
\label{tb:efficiency}
\begin{tabular}{cccc}
\toprule
Metric & Cost $\downarrow$ \\ \midrule
Model Parameters & 1.20M \\
Disk Storage & 9.76 MB\\
Memory Usage & 35MB\\
FLOPs & 233M \\
MAdd & 451M \\\bottomrule

\end{tabular}
\end{table}




\section{Conclusion}
In this paper, we have proposed an efficient framework for conducting federated learning on AIoT devices. Our approach significantly reduces the required computational cost and communication rounds which makes it suitable for AIoT devices. Through experiments on real-world datasets, we show that our method has a negligible loss in utility compared with the fully centralized counterpart. Hence, we believe that our approach is valuable for both the academia and industry.


\end{document}